\documentclass{article}
\usepackage{booktabs} 
\usepackage{authblk}
\usepackage{amsmath}
\usepackage{amsfonts}
\usepackage{amssymb}
\usepackage{graphicx}
\usepackage{hyperref}
\usepackage{float}
\usepackage{array} 
\usepackage{multirow}
\usepackage{listings}
\usepackage[top=3cm, bottom=3cm, left=2.5cm, right=2.5cm]{geometry} 
\usepackage[round, authoryear]{natbib} 
\usepackage{url} 

\newcolumntype{M}[1]{>{\centering\arraybackslash}m{#1}}

\begin{document}

\title{Leveraging Reasoning Model Answers to Enhance Non-Reasoning Model Capability}
\author{Haotian Wang}
\author{Han Zhao}
\author{Shuaiting Chen}
\author{Xiaoyu Tian}
\author{Sitong Zhao}
\author{Yunjie Ji}
\author{Yiping Peng}
\author{Xiangang Li}

\affil{
    \raisebox{-0.4em}{\includegraphics[height=1.5em]{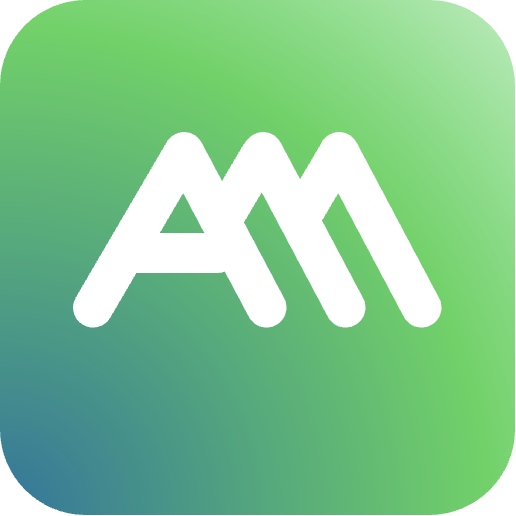}}
    \hspace{0.2em}a-m-team
}
\date{}

\maketitle

\begin{abstract}
Recent advancements in large language models (LLMs), such as DeepSeek-R1 and OpenAI-o1, have demonstrated the significant effectiveness of test-time scaling, achieving substantial performance gains across various benchmarks. These advanced models utilize deliberate "thinking" steps to systematically enhance answer quality. In this paper, we propose leveraging these high-quality outputs generated by reasoning-intensive models to improve less computationally demanding, non-reasoning models. We explore and compare methodologies for utilizing the answers produced by reasoning models to train and improve non-reasoning models. Through straightforward Supervised Fine-Tuning (SFT) experiments on established benchmarks, we demonstrate consistent improvements across various benchmarks, underscoring the potential of this approach
for advancing the ability of models to answer questions directly.
\end{abstract}

\begin{figure}[h!]
    \centering
    \includegraphics[width=0.75\linewidth]
    {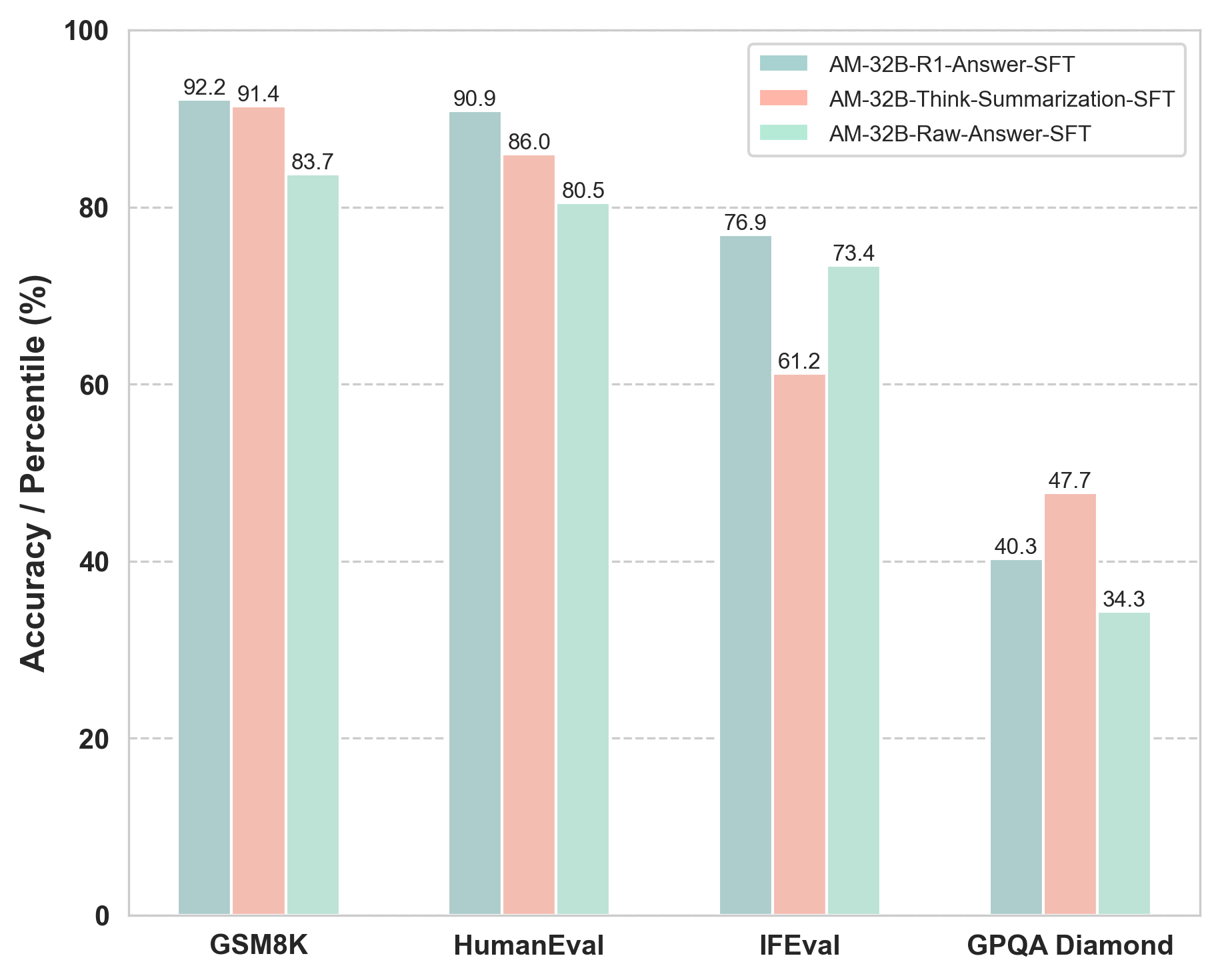}
    \caption{Benchmark performance of different answer utillize method}
    \label{fig:benchmar pic}
\end{figure}

\section{Introduction}

The field of Natural Language Processing has experienced remarkable advancements with the development of large language models (LLMs) \citep{min2021recentadvancesnaturallanguage, kaplan2020scalinglawsneurallanguage}. A significant trend in enhancing the capabilities of these models is test-time scaling \citep{yang2025thinkingoptimalscalingtesttimecompute, wu2025inferencescalinglawsempirical}, where increasing computational resources allocated during inference leads to notable performance improvements. Models such as OpenAI's o1 series \citep{OpenAI2024} and DeepSeek-R1 \citep{deepseekai2025deepseekr1incentivizingreasoningcapability} have demonstrated the effectiveness of this approach across various tasks and benchmarks \citep{lightman2023letsverifystepstep, Huang2024OpenCoderTO}. The capability of these models to achieve superior results by allocating additional computational resources during inference indicates an important shift in optimizing performance for LLMs. Specifically, dedicating more computation to the answer-generation process, rather than solely relying on scaling training data and model parameters, can lead to significant improvements, particularly in tasks that require complex reasoning \citep{snell2024scalingllmtesttimecompute}. The success of test-time scaling thus emphasizes the crucial role of computation during the answer-generation phase.

A key factor in the success of advanced models lies in their inherent ability to perform explicit reasoning before arriving at a final answer \citep{wei2023chainofthoughtpromptingelicitsreasoning, snell2024scalingllmtesttimecompute, wu2025inferencescalinglawsempirical}. This deliberate "think" step enables models to evaluate multiple potential solutions, resulting in more accurate and nuanced answers. Techniques like test-time scaling have facilitated the generation of these intermediate reasoning steps, significantly boosting performance on challenging tasks \citep{huang2022towards}. Given the enhanced quality of answers produced by such reasoning models, a relevant question emerges: Can these high-quality answers be effectively utilized to enhance the performance of less computationally intensive, non-reasoning models?

This paper investigates this question by exploring strategies for leveraging the outputs of reasoning models to enhance the capabilities of non-reasoning models. Our central hypothesis is that training non-reasoning models using improved answers derived from reasoning models can lead to superior performance. To validate this hypothesis, we conduct a series of experiments on several benchmark datasets, comparing different approaches to incorporating reasoning-model outputs into the training of non-reasoning models. Specifically, the contributions of this paper are: (1) we provide empirical evidence demonstrating that utilizing high-quality answers from reasoning models for training significantly enhances the performance of non-reasoning models. (2) we systematically investigate various methods for integrating outputs generated by reasoning models to improve the capabilities of non-reasoning models.

\section{Approach}

This section details our proposed approach, beginning with the generation of a well-distributed and diverse Supervised Fine-Tuning (SFT) dataset. Following this, we utilize DeepSeek-R1\citep{deepseekai2025deepseekr1incentivizingreasoningcapability}, yielding high-quality think and answer. We also utilize DeepSeek-v3-0324\citep{deepseekai2024deepseekv3technicalreport} to generate non-resoning response for comparison. Crucially we introduce and rigorously compare distinct methods for exploiting these distilled answers. The core of our technical contribution lies in proposing these varied utilization strategies and evaluating their effectiveness for improving non-reasoning models.

\subsection{Data Description}

Our dataset was constructed in two main stages: first, the collection of input prompts and second, the sampling of the corresponding responses.
\begin{itemize}
\item \textbf{Prompt Collection: }To ensure broad coverage and data diversity, we curated our dataset from open-source communities. It incorporates data derived from several established collections, including Infinity Instruct, OpenCoder\citep{Huang2024OpenCoderTO},PRIME\citep{cui2025process},NuminaMath\citep{numina_math_datasets},CodeContests\citep{doi:10.1126/science.abq1158}, FLAN\citep{weifinetuned}, Orca\citep{mukherjee2023orca}, AM-DeepSeek-R1-Distilled-1.4M\citep{AM-DeepSeek-R1-Distilled-1.4M},tuluv3\citep{lambert2024tulu3} and more. This aggregation spans multiple critical domains such as mathematics, code, science, general question answering, instruction following, tooluse and more.In total, our data collection efforts resulted in approximately 1.3 million instances.
\item \textbf{Response Collection: }Following the curation of our diverse query set, we utilized DeepSeek-R1 to generate reasoning process and answers.In addition to generating data via direct distillation from the DeepSeek-R1 model, we supplemented our dataset by directly selecting a subset of instances from the AM-DeepSeek-R1-Distilled-1.4M\citep{AM-DeepSeek-R1-Distilled-1.4M}, which represents one of the current state-of-the-art resources in this domain. To elicit high-quality and domain-appropriate outputs, we implemented a tailored prompting strategy. Specifically, distinct user prompts were designed and applied for queries falling within the mathematics and code domains, acknowledging their unique structural and logical demands compared to general tasks. Furthermore, to capture the reasoning process, the generated outputs were structured to distinguish between intermediate thinking steps (enclosed in \textless think\textgreater \textless /think\textgreater tags) and the final conclusive answer (enclosed in \textless answer\textgreater \textless /answer\textgreater  tags).  Detailed system prompts, domain-specific user prompts can be found in Appendix.
\end{itemize}
\begin{figure}[h!]
    \centering
    \includegraphics[width=0.75\linewidth]
    {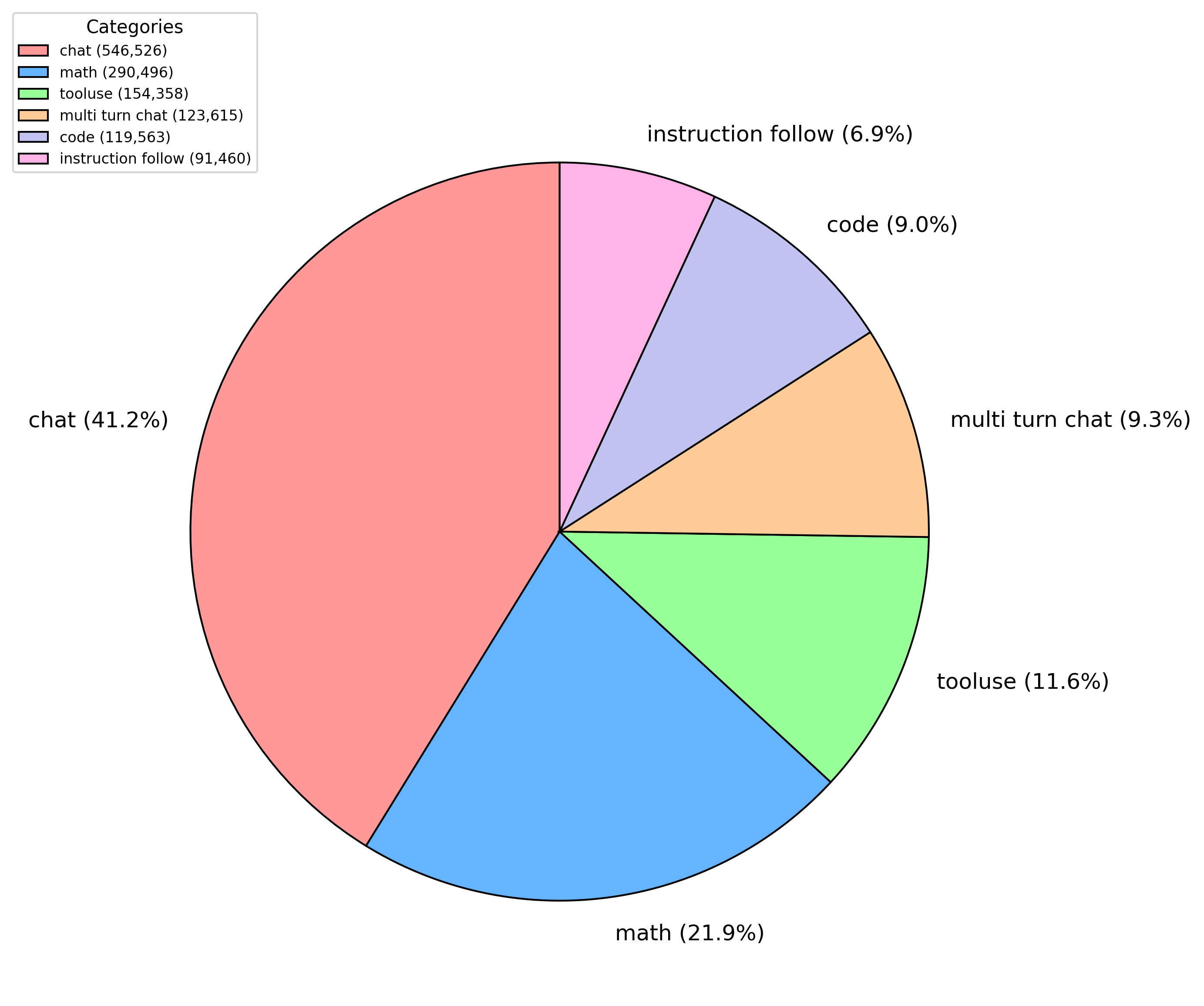}
    \caption{Category Distribution of dataset}
    \label{fig:category dist}
\end{figure}
\subsection{Methods for Utilizing Reasoning Response}

To evaluate different strategies for leveraging reasoning models to generate informative responses, we explored three distinct methods. Let $Q$ represent the input query. We define the following components:

\begin{itemize}
\item $R_{orig}$: The original response for query $Q$ obtained from the baseline open-source community dataset.
\item $M_{reason}$: The reasoning large language model.
\item ($T_{reason},A_{reason}$)=$M_{reason}(Q)$: The output of the reasoning model $M_{reason}$ for query $Q$,
consisting of an intermediate thinking component $T_{reason}$ and a final answer component$A_{reason}$.
\item $M_{sum}$: A separate summarization model (specifically, Qwen2.5-7B-Instruct) used in Method 4.
\item $\oplus$: An operator representing the concatenation of two text strings.
\end{itemize}
The three methods generate final responses, denoted $R_{1}$,$R_{2}$,$R_{3}$ respectively, as follows:

\begin{enumerate}
\item \textbf{Original Response: }This method utilizes the raw response directly from the community dataset, serving as our baseline for comparison. The final response $R_{1}$ is simply:
\begin{equation}
R_1 = R_{orig}
\label{eq:method1}
\end{equation}
This represents the unprocessed data from the source dataset.
\item \textbf{Direct Reasoning Model Output (Answer Component): }This approach uses only the answer component $A_{reason}$ generated directly by reasoning model $M_{reason}$.The final response $R_{2}$ is:
\begin{equation}
R_2 = A_{reason}
\label{eq:method2}
\end{equation}

This method isolates the direct answer produced by the reasoning model, potentially lacking the intermediate steps outlined in $T_{reason}$.

\item \textbf{Think Summarization: } To create a response that includes the reasoning process without excessive length, this method first summarizes the thinking component $T_{reason}$ using the summarization model $M_{sum}$. Let the generated summary be $S_{think}$, where:
\begin{equation}
S_{think} = M_{sum}(T_{reason})
\label{eq:metho34.1}
\end{equation}

This summary $S_{think}$ , capturing the essential problem-solving steps, is then prepended to the reasoning model's original answer component $A_{reason}$. The final response $R_{3}$ is constructed as :
\begin{equation}
R_3 = S_{think} \oplus A_{reason}
\label{eq:method3.2}
\end{equation}
The goal is to integrate the core reasoning path with the final answer, yielding a more comprehensive and explanatory output.
\end{enumerate}
In addition to the datasets generated by the three approaches outlined above, we utilized the DeepSeek-v3-0324\citep{deepseekai2024deepseekv3technicalreport} model to produce a fourth set of responses. This dataset serves as a control group for our analysis, bringing the total number of datasets evaluated in this paper to four.

\section{Experiments}
\subsection{Evaluation}
\subsubsection{Benchmark}
We evaluated non-reasoning model's ability using GPQA-Diamond \citep{rein2023gpqagraduatelevelgoogleproofqa}, GSM8K\citep{cobbe2021gsm8k},MMLU\citep{hendryckstest2021},HumanEval\citep{chen2021evaluating},IFEval\citep{zhou2023instructionfollowingevaluationlargelanguage},AlignBench\citep{liu2023alignbench}, MTBench\citep{zheng2023judging}. These benchmarks span multiple fields and difficulty levels, enabling a thorough assessment of the model's performance across diverse scenarios.

\subsubsection{Evaluation Methodology}To ensure consistent and comparable results across all evaluated models, we standardized the generation parameters. The maximum generation length was uniformly set to 16,384 tokens for all tasks.We employed two distinct decoding strategies based on the specific requirements of the evaluation benchmarks:

\begin{enumerate}
    \item \textbf{Stochastic Sampling:} For benchmarks requiring probabilistic generation, GPQA-Diamond \citep{rein2023gpqagraduatelevelgoogleproofqa}, we utilized a temperature of 0.6 and a top-p value of 0.95. To estimate the pass@1 metric for these benchmarks, we generated 8 candidate samples per input instance.
    \item \textbf{Greedy Decoding:} For all other benchmarks, we employed greedy decoding by setting the temperature to 0.0 and generating a single sample per input instance. This deterministic approach was applied to GSM8K \citep{cobbe2021gsm8k}, HumanEval \citep{chen2021evaluating}, IFEval \citep{zhou2023instructionfollowingevaluationlargelanguage}, MMLU \citep{hendryckstest2021}, AlignBench \citep{liu2023alignbench}, and MTBench \citep{zheng2023judging}.
\end{enumerate}
Within the greedy decoding setting, specific configurations were adopted for certain benchmarks:
\begin{itemize}
    \item IFEval\citep{zhou2023instructionfollowingevaluationlargelanguage}, we reported the prompt-strict score.
    \item For MMLU\citep{hendryckstest2021}, evaluations were conducted using a 5-shot prompting methodology.
    \item For AlignBench\citep{liu2023alignbench} and MTBench\citep{zheng2023judging}, the generated responses were subsequently evaluated using the OpenAI GPT-4 model \citep{OpenAI2024} as the judge.
\end{itemize}

\subsection{Experiment Setup}
We conducted SFT on Qwen2.5-32B\citep{qwen2.5} using datasets mentioned in Section 2.2. We employed a cosine
Learning Rate Scheduler, set the Learning Rate to 8e-6, the Warm Up Ratio to 0.05, the Batch Size to 64,
and the Max Token Length to 16,384. The training process consisted of three epochs.


\subsection{Results and Analysis}

\begin{table}[htbp]
  \caption{SFT-Model Performance}
  \vspace{0.5em}
  \centering
  \renewcommand{\arraystretch}{1.5}
  \label{tab:model_performance}
  \resizebox{\textwidth}{!}{
    \begin{tabular}{l||
                      M{\dimexpr0.16\textwidth\relax}
                      M{\dimexpr0.16\textwidth\relax}
                      M{\dimexpr0.10\textwidth\relax}
                      M{\dimexpr0.10\textwidth\relax}
                      M{\dimexpr0.12\textwidth\relax}
                      M{\dimexpr0.17\textwidth\relax}
                      M{\dimexpr0.12\textwidth\relax}
                      M{\dimexpr0.12\textwidth\relax}}
      \hline
      \textbf{Model} & \textbf{GPQA-Diamond pass@1} & \textbf{GSM8K} & \textbf{MMLU} & \textbf{HumanEval} & \textbf{IFEval \ \ \ (prompt strict)} & \textbf{AlignBench} & \textbf{MTBench} \\[2ex]
      \hline
      \textbf{OLMo-2-32B-0325-SFT} & \textbf{-} & 78.4 & 76.1 & - & 72.4 & \textbf{-} & \textbf{-} \\
      \textbf{AM-32B-DeepSeek-V3-Answer-SFT} & \textbf{47.9} & 90.2 & 70.1 & 90.9 & 76.6 & \textbf{7.6} & \textbf{8.2} \\
      \hline

      \textbf{AM-32B-R1-Answer-SFT} & 40.3 & \textbf{92.2} & \textbf{82.5} & \textbf{90.9} & 76.9 & 6.9 & 7.6 \\
      \textbf{AM-32B-Think-Summarization-SFT} & 47.7 & 91.4 & 81.0 & 86.0 & 61.2 & 7.2 & 7.9 \\
        \hline
          \textbf{AM-32B-Raw-Answer-SFT(Baseline)} & 34.3 & 83.7 & 82.5 & 80.5 & 73.4 & 6.2 & 7.7 \\
          \hline
    \end{tabular}
  }
\end{table}

Our experimental results demonstrate that directly using the answer portion from response of reasoning model for model training via Supervised Fine-Tuning (SFT) leads to significant improvements on several key benchmarks, particularly HumanEval\citep{chen2021evaluating}, GSM8K\citep{cobbe2021gsm8k}, and GPQA\citep{rein2023gpqagraduatelevelgoogleproofqa}. However, we observed a slight decrease in performance on chat-oriented metrics such as AlignBench\citep{liu2023alignbench} and MT Bench\citep{zheng2023judging}. Our analysis focuses on elucidating the impact of different data generation methods derived from a reasoning model's output on the fine-tuned non-reasoning model's capabilities across diverse benchmarks.
\textbf{Baseline Performance:} The AM-32B-Raw-Answer-SFT model, fine-tuned directly on the original community dataset responses ($R_{orig}$), serves as our primary baseline. It establishes a reference point for evaluating the efficacy of incorporating reasoning model outputs.

\textbf{Impact of Direct Reasoning Answers:} Utilizing only the direct answer component ($A_{reason}$) from the reasoning model for SFT (AM-32B-R1-Answer-SFT) yielded substantial performance improvements on several key reasoning and coding benchmarks. As indicated in Table \ref{tab:model_performance}, this model achieved the highest scores among our SFT variants on GSM8K\citep{cobbe2021gsm8k} (92.2) and HumanEval\citep{chen2021evaluating} (90.9), and demonstrated significant gains on GPQA-Diamond \citep{rein2023gpqagraduatelevelgoogleproofqa} (40.3) compared to the baseline (34.3). However, this approach resulted in marginally lower performance on chat-oriented benchmarks, namely AlignBench\citep{liu2023alignbench} and MTBench\citep{zheng2023judging}, compared to the baseline. We attribute this phenomenon to the structure of the reasoning model's output; the detailed procedural explanations often reside within the reasoning trace ($T_{reason}$), leaving the answer component ($A_{reason}$) overly concise and potentially lacking the conversational context preferred in chat interactions.

\textbf{Benefits and Trade-offs of Think Summarization}: The AM-32B-Think-Summary-SFT model, trained using the summarized thinking process concatenated with the answer ($S_{think}$ $\oplus$
$A_{reason}$), was designed to mitigate the conversational shortcomings observed with the Think Summarization approach. This method achieved the highest GPQA\citep{rein2023gpqagraduatelevelgoogleproofqa} score (47.7) among our models, approaching the performance of the distilled external model, and improved performance on MTBench\citep{zheng2023judging} (7.9). These results suggest that explicitly incorporating a summary of the reasoning process enhances understanding and potentially improves alignment in conversational contexts. However, this structural modification introduced a notable trade-off, evidenced by a significant decrease in performance on IFEval\citep{zhou2023instructionfollowingevaluationlargelanguage} (61.2). We find that the substantial alteration of the original answer format may interfere with the model's ability to adhere strictly to instructions as defined by the IFEval\citep{zhou2023instructionfollowingevaluationlargelanguage} benchmark. Despite this specific deficit, the Think Summarization method generally provided a more balanced performance profile across reasoning and chat-related tasks compared to using the direct answer alone.

\textbf{Implications for Utilize Reasoning Model's Power:} Comparing our results, particularly the performance of AM-32B-R1-Answer-SFT, with the externally sourced AM-32B-DeepSeek-V3-Answer-SFT model highlights a critical insight. Simply fine-tuning on the final answers ($A_{reason}$) extracted from a capable reasoning model, while beneficial for certain tasks, does not automatically transfer the full spectrum of the source model's capabilities, especially concerning conversational abilities or potentially nuanced reasoning reflected in the thought process. This underscores the necessity of developing more sophisticated methods that effectively structure and integrate both the reasoning process and the final answer when the objective is knowledge distillation or holistic capability transfer via SFT. The effectiveness of the Think Summarization approach, despite its IFEval limitation, points towards the importance of response content organization in this process.

\section{Discussion and Conclusion}
The results presented in this paper affirm that supervised fine-tuning (SFT) using response data derived from reasoning models can significantly enhance the performance of target language models. Our investigation systematically evaluated three distinct methodologies for utilizing these reasoning-derived outputs, revealing that the effectiveness of knowledge transfer is critically dependent on the specific strategy employed for structuring the SFT data.

Crucially, our findings demonstrate that simply leveraging the final answer component ($A_{reason}$) from a reasoning process, while boosting performance on certain reasoning and coding benchmarks, may not yield holistic improvements and can even slightly degrade performance in conversational alignment metrics. This highlights the importance of the information's structure; methods incorporating summarized reasoning steps ($S_{think}$) offered alternative performance profiles, often achieving better balance across diverse tasks or excelling in specific areas like instruction following, albeit sometimes involving trade-offs (e.g., the observed IFEval performance reduction with the Think Summarization method).

These results underscore the potential of leveraging reasoning outputs as a potent form of data augmentation for SFT, offering a viable pathway towards enhancing the capabilities of large language models. The variations in performance across methods emphasize that the manner in which reasoning-derived knowledge is structured and presented during fine-tuning is a key determinant of the resulting model's strengths and weaknesses. This work contributes practical strategies for such capability transfer, demonstrating tangible improvements across multiple standard benchmarks.

Building on these findings, future research should explore more sophisticated techniques for extracting, representing, and integrating the knowledge embedded within the reasoning process ($T_{reason}$). Investigating alternative summarization strategies, methods for dynamically combining reasoning steps with final answers, or techniques for explicitly modeling the reasoning structure could potentially unlock further performance gains and lead to the development of more robust and versatile models through optimized knowledge distillation.

\section{Limitation}
A notable limitation of the present study pertains to the Think Summarization methodology employed for generating training data. While this approach successfully integrates aspects of the reasoning process by summarizing the thinking trace ($T_{reason}$), the resulting summarized steps ($S_{think}$) may not fully capture the original fidelity or granularity inherent in the reasoning model's detailed thought process. Consequently, the SFT data generated via this method might represent a less precise approximation of the underlying reasoning compared to the original trace itself.

An alternative strategy, warranting further investigation, involves leveraging prompt engineering. Specifically, one could potentially refine the prompts used to elicit responses from the source reasoning model, instructing it to directly incorporate essential, concise reasoning steps within the structure of its final answer. This approach could obviate the need for post-hoc summarization and potentially yield training data that more faithfully represents integrated reasoning and conclusions.

However, the systematic exploration and optimization of such prompt engineering techniques fell outside the defined scope of the current project due to practical constraints on time and resources. Nevertheless, developing methods to elicit more integrated reasoning directly via prompting remains a promising avenue for future research, potentially offering a more direct and higher-fidelity approach to creating SFT data that effectively transfers reasoning capabilities.

\bibliographystyle{plainnat}
\bibliography{references}

\clearpage
\appendix

\section{Prompt}
\label{a}
\begin{figure}[h!]
    \centering
    \includegraphics[width=0.75\linewidth]
    {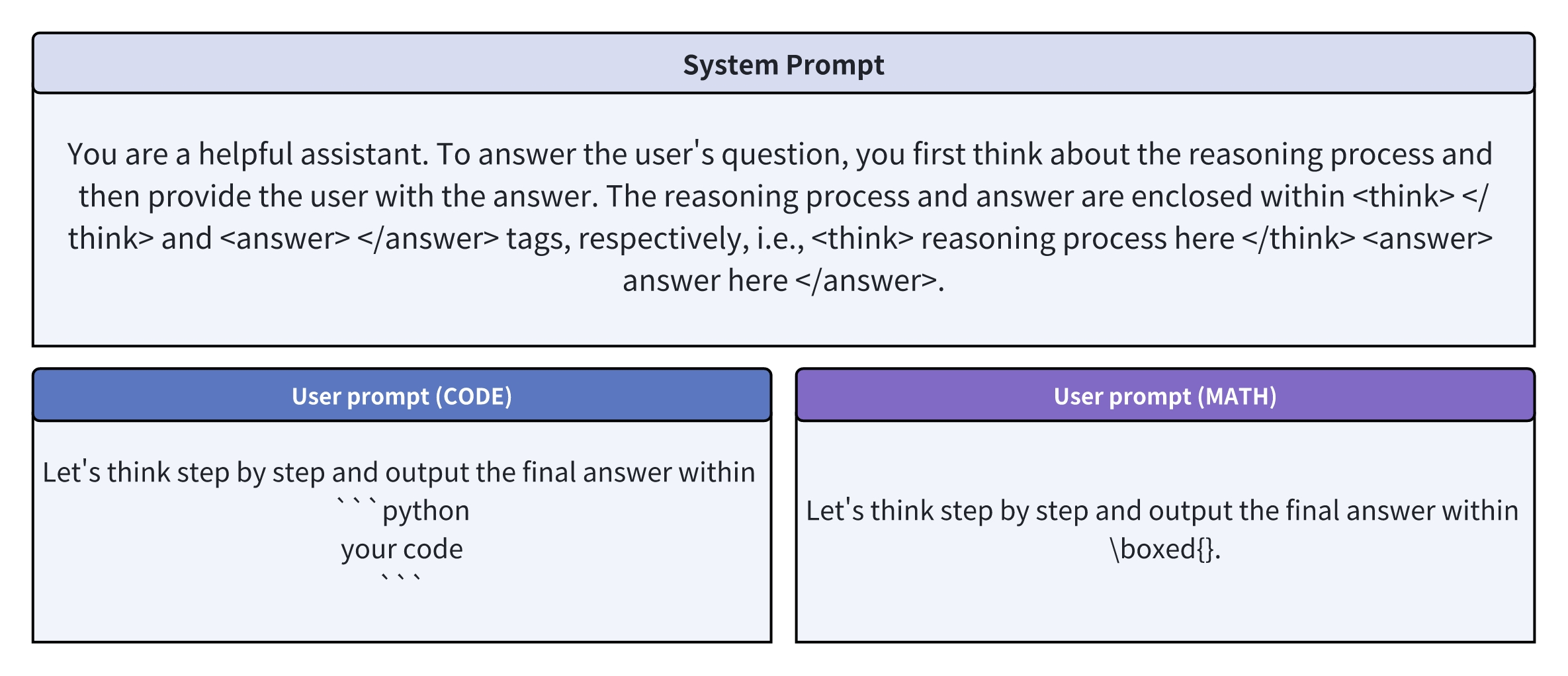}
    \caption{prompt used in inference}
    \label{fig:prompt used}
\end{figure}

\end{document}